\documentclass[a4paper,12pt]{article}

\usepackage{comment}
\usepackage{fancyvrb}
\usepackage{graphicx}
\usepackage{hyperref}
\usepackage[ragged]{footmisc}
\usepackage{ragged2e}
\usepackage[section]{placeins}

\begin{document}

\title{Solutions to problems with deep learning}

\author{J Gerard Wolff\footnote{Dr Gerry Wolff, BA (Cantab), PhD (Wales), CEng, MBCS, MIEEE; CognitionResearch.org, Menai Bridge, UK; \href{mailto:jgw@cognitionresearch.org}{jgw@cognitionresearch.org}; +44 (0) 1248 712962; +44 (0) 7746 290775; {\em Skype}: gerry.wolff; {\em Web}: \href{http://www.cognitionresearch.org}{www.cognitionresearch.org}.}}

\maketitle

\begin{abstract}

Despite the several successes of deep learning systems, there are concerns about their limitations, discussed most recently by Gary Marcus. This paper discusses Marcus's concerns and some others, together with solutions to several of these problems provided by the {\em SP theory of intelligence} and its realisation in the {\em SP computer model}. The main advantages of the SP system are: relatively small requirements for data and the ability to learn from a single experience; the ability to model both hierarchical and non-hierarchical structures; strengths in several kinds of reasoning, including `commonsense' reasoning; transparency in the representation of knowledge, and the provision of an audit trail for all processing; the likelihood that the SP system could not be fooled into bizarre or eccentric recognition of stimuli, as deep learning systems can be; the SP system provides a robust solution to the problem of `catastrophic forgetting' in deep learning systems; the SP system provides a theoretically-coherent solution to the problems of correcting over- and under-generalisations in learning, and learning correct structures despite errors in data; unlike most research on deep learning, the SP programme of research draws extensively on research on human learning, perception, and cognition; and the SP programme of research has an overarching theory, supported by evidence, something that is largely missing from research on deep learning. In general, the SP system provides a much firmer foundation than deep learning for the development of artificial general intelligence.

\end{abstract}

\section{Introduction}

Deep learning has received a great deal of attention, largely because of what it can do well, but there are concerns about its limitations, discussed most recently by Gary Marcus \cite{marcus_2018}.

This short paper discusses Marcus's concerns briefly and some others, together with solutions to several of these problems provided by the {\em SP theory of intelligence} and its realisation in the {\em SP computer model} \cite{sp_extended_overview,wolff_2006}, outlined in \cite[Appendix A]{sp_ghlai_2017} with pointers to where fuller information may be found.\footnote{Publications in the SP programme of research, most with download links, may be found on \href{http://www.cognitionresearch.org/sp.htm}{www.cognitionresearch.org/sp.htm}.}

Several of these solutions have been described in \cite[Section V]{sp_alternatives}. References to this paper are made at appropriate points below.

\section{What deep learning does well}

As Marcus says: ``Deep learning, as it is primarily used, is essentially a statistical technique for classifying
patterns, based on sample data, using neural networks with multiple layers.'' \cite[p.~3]{marcus_2018}. Even in applications such as the playing of games, it seems that the primary function of deep learning is in the recognition of patterns.

\section{Problems with deep learning and how they may be solved}

In this section, the first 10 subheadings are the same as in \cite{marcus_2018}, with the relevant section number shown at the end of each heading. The remaining headings are drawn largely from \cite[Section V]{sp_alternatives}, unless they are already covered by the first 10 headings.

\subsection{Deep learning thus far is data hungry (3.1)}\label{data_hungry_section}



The data-hungry nature of deep learning and how this may be overcome in the SP system has been discussed quite fully in \cite[Section V-E]{sp_alternatives}. There is also relevant discussion in \cite[Section V-D]{sp_alternatives}.

In brief, the SP system, like a person, can learn from a single exposure or experience, and it can begin to form meaningful new structures and generalisations with exposure to a mere handful of other examples.

In this connection, the SP system is much more like a child than are deep learning systems: neuroscientist David Cox has been reported as saying: ``To build a dog detector [with a deep learning system], you need to show the program thousands of things that are dogs and thousands that aren't dogs. My daughter only had to see one dog.'' and, the report says, she was happily pointing out puppies ever since.\footnote{``Inside the moonshot effort to finally figure out the brain'', {\em MIT Technology Review}, 2017-10-12, \href{http://bit.ly/2wRxsOg}{bit.ly/2wRxsOg}.}

What about the slow learning of complex skills like speaking and understanding a new language or how to play a piano? With deep learning, it is assumed that this kind of slow acquisition of skills may be explained by the gradual strengthening of links in Hebbian-style learning. By contrast, in the SP system, the slow learning of complex skills may be explained by the complexity of the search that is required to find good structures.

In summary, the SP system can explain learning from a single exposure or experience and it can explain the slow learning of complex skills. By contrast, a deep learning system can explain the slow learning of complex skills, but it fails to explain how learning may be achieved with a single exposure or experience, and more generally, it is excessively demanding in its requirements for data.

\subsection{Deep learning thus far is shallow and has limited capacity for
transfer (3.2)}\label{shallow_limited_transfer_section}

In \cite[Section 3.2]{marcus_2018}, Marcus points out quite rightly that ``it is important to realize that the word `deep' in deep learning refers to a technical, architectural property (the large number of hidden layers used in a modern neural networks, where there predecessors used only one)'' (p.~7).

He goes on to say that it is easy to over-interpret the results from a deep learning system. For example, ``according to a widely-circulated video of the system learning to play the brick-breaking Atari game Breakout, `after 240 minutes of training, [the system] realizes that digging a tunnel through the wall is the most effective technique to beat the game'. But the system has learned no such thing; it doesn't really understand what a tunnel, or what a wall is; it has just learned specific contingencies for particular scenarios. Transfer tests---in which the deep reinforcement learning system is confronted with scenarios that differ in minor ways from the one ones on which the system was trained show that deep reinforcement learning’s solutions are often extremely superficial.'' (p.~8).

The SP system certainly does not provide a comprehensive solution to issues like those just described. In brief, it seems fair to summarise the strengths and potential of the SP system, and to compare it with deep learning systems, as follows:

\begin{itemize}

    \item The SP system has strengths in the representation of diverse kinds of knowledge, in diverse aspects of intelligence, and in the seamless integration of diverse kinds of knowledge and diverse aspects of intelligence, in any combination. These strengths, which all flow from the powerful concept of {\em SP-multiple-alignment}, are summarised in \cite[Sections 3, 4, and 5]{sp_ghlai_2017}, with pointers to where fuller information may be found.

    \item Although the SP system has strengths in the representation of diverse kinds of knowledge, it seems likely that more research will be required to understand how the system may learn and represent the great range of concepts employed by people. There is some discussion in \cite[Sections 6.1 and 6.2]{sp_vision} about how the system may develop a concept of a three-dimensional object, and in \cite[Section 5.3]{sp_vision}, there is brief discussion of the development of concepts like motion and speed.

    \item At a `deep' level, it seems likely that {\em all} kinds of learning, both in deep learning systems and in the SP system, may be understood as the learning of statistical contingencies.

\end{itemize}

\subsection{Deep learning thus far has no natural way to deal with
hierarchical structure (3.3)}\label{problems_with_hierarchical_structure_section}

Computer models developed in a programme of research on language learning (summarised in \cite{sp_extended_overview}) were designed to work by `hierarchical chunking'. As one might expect, these models were good at representing hierarchical structures.

But in the `SP' programme of research---where the aim has been to simplify and integrate observations and concepts across artificial intelligence, mainstream computing, mathematics, and human learning, perception, and cognition---hierarchical chunking would not do. The challenge has been to create a framework that would serve equally well for the representation of both hierarchical and non-hierarchical structures.

\begin{figure}[!htbp]
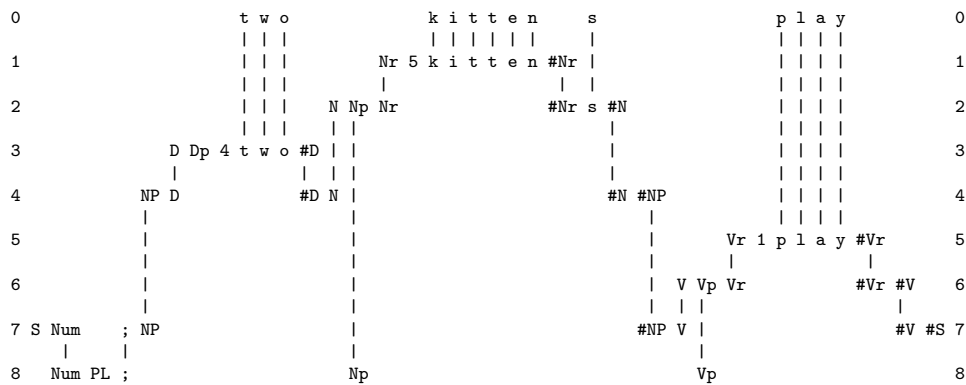

\fontsize{07.00pt}{08.40pt}
\centering
{\bf
\begin{BVerbatim}
0                      t w o              k i t t e n     s                  p l a y           0
                       | | |              | | | | | |     |                  | | | |
1                      | | |         Nr 5 k i t t e n #Nr |                  | | | |           1
                       | | |         |                 |  |                  | | | |
2                      | | |    N Np Nr               #Nr s #N               | | | |           2
                       | | |    | |                         |                | | | |
3               D Dp 4 t w o #D | |                         |                | | | |           3
                |            |  | |                         |                | | | |
4            NP D            #D N |                         #N #NP           | | | |           4
             |                    |                             |            | | | |
5            |                    |                             |       Vr 1 p l a y #Vr       5
             |                    |                             |       |             |
6            |                    |                             |  V Vp Vr           #Vr #V    6
             |                    |                             |  | |                   |
7 S Num    ; NP                   |                            #NP V |                   #V #S 7
     |     |                      |                                  |
8   Num PL ;                      Np                                 Vp                        8
\end{BVerbatim}
}
\caption{The best SP-multiple-alignment created by the SP computer model with a store of SP-patterns like those in rows 1 to 8 (representing grammatical structures, including words) and an SP-pattern representing a sentence to be parsed shown in row 0.}
\label{parsing_figure}
\end{figure}

\sloppy What has proved to be a good solution is the concept of {\em SP-multiple-alignment}, borrowed and adapted from the concept of `multiple sequence alignment' in bioinformatics. Although the basic idea is to create alignments of two or more sequences,\footnote{It is envisaged that at some stage the SP system will be adapted work with two-dimensional patterns as well as on-dimensional sequences.} the framework lends itself very well to the representation of the kinds of hierarchical structures recognised in linguistic analysis, as can be seen in Figure \ref{parsing_figure}.

There is more discussion in \cite[Section V-K]{sp_alternatives}.


\subsection{Deep learning thus far has struggled with open-ended inference (3.4)}\label{open_ended_inference_section}

No work has yet been done to explore whether or how the SP system can model `open ended' inferences but, unlike deep learning systems, it has strengths in several different forms of reasoning including: one-step `deductive' reasoning; chains of reasoning; abductive reasoning; reasoning with probabilistic networks and trees; reasoning with `rules'; nonmonotonic reasoning and reasoning with default values; Bayesian reasoning with `explaining away'; causal reasoning; reasoning that is not supported by evidence; the inheritance of attributes in class hierarchies; and inheritance of contexts in part-whole hierarchies (\cite[Chapter 7]{wolff_2006}, \cite[Section 10]{sp_extended_overview}). Where it is appropriate, probabilities for inferences may be calculated in a straightforward manner (\cite[Section 3.7]{wolff_2006}, \cite[Section 4.4]{sp_extended_overview}). There is also potential for spatial reasoning \cite[Section V-F.1]{sp_autonomous_robots}, and for what-if reasoning \cite[Section V-F.2]{sp_autonomous_robots}.

There is more discussion in \cite[Section V-L]{sp_alternatives}.

\subsection{Deep learning thus far is not sufficiently transparent (3.5)}\label{transparency_section}

Marcus \cite[Section 3.5]{marcus_2018} writes that ``Although some strides have been [made] in visualizing the contributions of individuals nodes in complex networks ..., most observers would acknowledge that neural networks as a whole remain something of a black box.'' (p.~11).

In this respect, there is a sharp contrast with the SP system \cite[Section V-J]{sp_alternatives}:

\begin{itemize}

    \item All knowledge stored by the SP system is transparent and open to inspection.

    \item In general, knowledge in the SP system is likely to be comprehensible by people but problems may arise with concepts that have not yet been well studied in the SP programme of research (Section \ref{shallow_limited_transfer_section}).

    \item There is an audit trail for all processing performed by the SP system and all conclusions it may reach.

\end{itemize}


\subsection{Deep learning thus far has not been well integrated with prior
knowledge (3.6)}\label{not_well_integrated_with_prior_knowledge_section}

Marcus \cite[Section 3.6]{marcus_2018} writes that ``Work in deep learning typically consists of finding a training database, sets of inputs associated with respective outputs, and learn all that is required for the problem by learning the relations between those inputs and outputs, using whatever clever architectural variants one might devise, along with techniques for cleaning and augmenting the data set. With just a handful of exceptions, ..., prior knowledge is often deliberately minimized.'' (p.~11).

Later, he writes that ``It also not straightforward in general how to integrate prior knowledge into a deep
learning system:, in part because the knowledge represented in deep learning systems pertains mainly to (largely opaque) correlations between features, rather than to abstractions like quantified statements (e.g. all men are mortal), see discussion of universally-quantified one-to-one-mappings in Marcus (2001), or generics ....'' (p.~11).

Similar things may be said about the SP system but, here, the problems are likely to be less severe. This is because, in general, prior knowledge may be represented in the same format as knowledge that the system learns for itself.

Later in the same section, Marcus writes:

\begin{quote}

    ``Problems that have less to do with categorization and more to do with commonsense reasoning essentially lie outside the scope of what deep learning is appropriate for, and so far as I can tell, deep learning has little to offer such problems. In a recent review of commonsense reasoning, Ernie Davis and I \cite{davis_marcus_2015} began with a set of easily-drawn inferences that people can readily answer without anything like direct training, such as {\em Who is taller, Prince William or his baby son Prince George? Can you make a salad out of a polyester shirt? If you stick a pin into a carrot, does it make a hole in the carrot or in the pin?}

    ``As far as I know, nobody has even tried to tackle this sort of thing with deep learning.

    ``Such apparently simple problems require humans to integrate knowledge across vastly disparate sources, and as such are a long way from the sweet spot of deep learning-style perceptual classification. Instead, they are perhaps best thought of as a sign that entirely different sorts of tools are needed, along with deep learning, if we are to reach human-level cognitive flexibility.'' (p.~12).

\end{quote}

It appears that the SP system has considerable potential with the kinds of commonsense reasoning discussed by Davis and Marcus \cite{davis_marcus_2015}. This is described in \cite{sp_csrk}, a detailed response to the issues raised in their paper.

\subsection{Deep learning thus far cannot inherently distinguish causation
from correlation (3.7)}\label{causation_correlation_section}

Marcus \cite[Section 3.7]{marcus_2018} writes: ``If it is a truism that causation does not equal correlation, the distinction between the two is also a serious concern for deep learning. Roughly speaking, deep learning learns
complex correlations between input and output features, but with no inherent representation of causality.'' (p.~12--13).

It cannot be claimed that there is a comprehensive analysis, within the SP system, of the difference between causation and correlation. But the system does produce useful demonstrations of how such concepts may be modelled in the system:

\begin{itemize}

    \item {\em Causation}. In \cite[Section 10.5]{sp_extended_overview} and \cite[Section 7.9]{wolff_2006}, there is a demonstration of how the SP system may serve in the causal diagnosis of faults in an electronic circuit.

    \item {\em Correlation}. In \cite[Section 10.2]{sp_extended_overview} and \cite[Section 7.8]{wolff_2006}, there is an example showing how Bayesian reasoning, with conditional probabilities centre stage, may be modelled in the SP system.

\end{itemize}

\subsection{Deep learning presumes a largely stable world, in ways that may
be problematic (3.8)}\label{stable_world_assumption_section}

Marcus \cite[Section 3.8]{marcus_2018} writes: ``The logic of deep learning is such that it is likely to work best in highly stable worlds, like the board game Go, which has unvarying rules, and less well in systems such as politics and economics that are constantly changing.'' (p.~13).

Much the same may be said of the SP system, or any other system that models the world via its statistical structure, which is the rule for most AI systems.

The trick, of course, for people and for artificial systems, is to be prepared for constant changes in the phenomena that one is modelling. And with predictions in areas such as economics, there is the possibility that each prediction itself may change what it is that is being predicted!

\subsection{Deep learning thus far works well as an approximation, but its
answers often cannot be fully trusted (3.9)}\label{approximation_and_trust_section}

Marcus \cite[Section 3.9]{marcus_2018} notes that there are now many examples where deep learning systems are fooled into making misclassifications which appear eccentric or bizarre to people. In \cite[Section V-G]{sp_alternatives} I have described similar examples.

In the latter section I have written:

\begin{quote}

    ``With regard to the first kind of error---failing to recognise something that is almost identical to what has been recognised---there is already evidence that the SP computer model would not make that kind of mistake. It can recognise words containing errors of omission, commission and substitution \cite[Section 6.2.1]{wolff_2006}, and likewise for diseases in medical diagnosis viewed as pattern recognition \cite[Section 3.6]{wolff_medical_diagnosis} and in the parsing of natural language [2, Section 4.2.2].

    ``No attempt has been made to test experimentally whether or not the SP computer model is prone to the second kind of error---recognising abstract patterns as ordinary objects---but a knowledge of how it works suggests that it would not be.''

\end{quote}

In general, it seems that the SP system is unlikely to make the strange errors of classification to which deep learning systems are prone.

\subsection{Deep learning thus far is difficult to engineer with (3.10)}\label{difficult_to_engineer_with_section}

Marcus \cite[Section 3.10]{marcus_2018} writes:

\begin{quote}

    ``Another fact that follows from all the issues raised above is that is simply hard to do robust engineering with deep learning. As a team of authors at Google put it in 2014, in the title of an important, and as yet unanswered essay \cite{sculley_etal_2014}, machine learning is `the high-interest credit card of technical debt', meaning that is comparatively easy to make systems that work in some limited set of circumstances (short term gain), but quite difficult to guarantee that they will work in alternative circumstances with novel data that may not resemble previous training data (long term debt, particularly if one system is used as an element in another larger system).

    ``In an important talk at ICML, Leon Bottou \cite{bottou_2015} compared machine learning to the development of an airplane engine, and noted that while the airplane design relies on building complex systems out of simpler systems for which it was possible to create sound guarantees about performance, machine learning lacks the capacity to produce comparable guarantees. As Google's Peter Norvig \cite{norvig_2016} has noted, machine learning as yet lacks the incrementality, transparency and debuggability of classical programming, trading off a kind of simplicity for deep challenges in achieving robustness.

    ``Henderson and colleagues have recently extended these points, with a focus on deep reinforcement learning, noting some serious issues in the field related to robustness and replicability \cite{henderson_2017}.'' (p.~14).

\end{quote}

Evidence to date suggests that these remarks are unlikely to apply to the SP system. Although development of the system has taken a long time, the SP computer model now exhibits considerable stability and robustness, and promises to provide a sound basis for scaling up with parallel processing, and for further developments as noted in \cite[Section 3.3]{sp_extended_overview}.

\subsection{Catastrophic forgetting}\label{catastrophic_forgetting_section}

A weakness of deep learning that was overlooked in the writing of \cite{sp_alternatives} is a problem called ``catastrophic forgetting'', meaning the way in which new learning in a deep learning system wipes out old memories. A solution has been proposed in \cite{kirkpatrick_2017} but it appears to be partial, and it is unlikely to be satisfactory in the long run.

The SP system is immune to any such influence in its learning. All learning is achieved by the addition of new SP-patterns to the system's store of SP-patterns, and, while there may be some merging of new information with old information that is similar, new learning does not disturb old learning.

Of course, there may be a case for introducing some kind of forgetting into the system, but the system as it is now does not forget.

\subsection{Under- and over-generalisations and their correction}\label{under-over-generalisation_section}


A general problem in any kind of learning system, especially the learning of a natural language, is how to generalise `correctly' from the finite sample of information which is the basis for learning---avoiding both over-generalisations and under-generalisations. A related problem is learning from `dirty data': how to learn `correct' structures despite the fact that most samples of data contain `errors'---and this in the face of evidence that learning may be successful without the opportunity for children to receive correction from adults or older children.

Several solutions have been proposed for deep learning systems, some of which are referenced in \cite[Section V-N]{sp_alternatives}. But I believe it is fair to say that none of them are entirely satisfactory.

What I believe is a much better solution has been described in \cite[Section 5.3]{sp_extended_overview} and in \cite[Section V-N]{sp_alternatives}. In brief:

\begin{enumerate}

    \item Given a finite sample of data, {\bf I}, compress it as much as possible to yield a {\em grammar}, {\bf G}, and an {\em encoding}, {\bf E}, of {\bf I} in terms of {\bf G}.

    \item In general, it will be found that {\bf G} represents the `essence' of {\bf I}, without over- or under-generalisations, and without `errors' from the `dirty data'.

\end{enumerate}

Naturally, there are many associated issues that may be discussed but I believe that the framework outlined here will prove to be sound.

\subsection{Psychology and neuroscience}\label{psychology_neuroscience_section}

It has long been recognised that deep learning systems are only loosely related to any kind of structure or processing in the brain. More generally, research on deep learning has been conducted largely without reference to what has been learned about human learning, perception, and cognition, or what is known about neuroscience.

By contrast, the SP system draws extensively on my own background in cognitive psychology, and my extensive programme of research on the learning of a first language or languages by children (summarised in \cite{wolff_1988}).

From its beginnings, the SP programme of research has been influenced by a long-running theme, beginning with research by Fred Attneave \cite{attneave_1954}, Horace Barlow \cite{barlow_1959,barlow_1969} and others, pointing to the importance of information compression in human learning, perception, and cognition.

It is also relevant to mention that abstract structures and processes in the SP system, map quite neatly on to what appear to be plausible structures of neurons and their interconnections, and how they may function, described in \cite{spneural_2016}.

\subsection{Overarching theory}\label{overarching_theory_section}

A variety of sources, perhaps most notably the work of Ray Solomonoff \cite{solomonoff_1964,solomonoff_1997} point to the importance of information compression in learning.

In research on deep learning in artificial neural networks, well reviewed by J{\"u}rgen Schmidhuber \cite{schmidhuber_2015}, there is some recognition of the importance of information compression \cite[Sections 4.2, 4.4, and 5.6.3]{schmidhuber_2015}, but it appears that the idea is not well developed in deep learning systems.

By contrast, as readers may have learned from \cite{sp_extended_overview,wolff_2006}, and may guess from Section \ref{under-over-generalisation_section}, the SP system is devoted to the compression of information, and more precisely compression of information via the powerful concept of SP-multiple-alignment, illustrated in Figure \ref{parsing_figure}. There is much evidence in support of this theory presented in \cite{sp_extended_overview,wolff_2006} and elsewhere.

In general, the SP system has a coherent overarching theory, something which is largely missing from research in deep learning.

\section{Potential risks of excessive hype}\label{risks_of_hype_section}

In \cite[Section 4]{marcus_2018}, Marcus writes:

\begin{quote}

    ``My own largest fear is that the field of AI could get trapped in a local minimum, dwelling too heavily in the wrong part of intellectual space, focusing too much on the detailed exploration of a particular class of accessible but limited models that are geared around capturing low-hanging fruit---potentially neglecting riskier excursions that might ultimately lead to a more robust path.''

\end{quote}

This chimes very much with my own experience. Despite many respectable publications in the SP programme of research, and many useful results, it has proved very difficult to get a hearing for this work by those engaged in research on deep learning. It seems that the extraordinary enthusiasm for deep learning, perhaps coupled with the large amounts of money being channelled into this area, has made it very difficult for researchers to divert any of their attention to anything other than deep learning. One senior researcher, who was kind enough to reply to one of my emails, said that although the SP research may be interesting, he does not have the time to look at it.

Good science and engineering is not like this. To solve difficult problems it is necessary to maintain several paths through the search space, and for researchers in any one area to be prepared to keep abreast of developments in other areas. In keeping with that approach, the central aim of the SP programme of research is simplification and integration of observations and concepts across artificial intelligence, mainstream computing, mathematics, and human learning, perception, and cognition.

Overspecialisation is a phenomenon noted by John Kelly and Steve Hamm, both of IBM, in their book {\em Smart Machines} \cite{kelly_hamm_2013}. In connection with research on different sensory modalities they write:

\begin{quote}

    ``In order to make this great leap and become true thinking machines, the cognitive systems of the future will integrate information from multiple sensing technologies. Today, as scientists labor to create machine technologies to augment our senses, there's a strong tendency to view each sensory field in isolation as specialists focus only on a single sensory capability. Experts in each sense don't read journals devoted to the others senses, and they don't attend one another's conferences. Even within IBM, our specialists in different sensing technologies don't interact much. Yet if machines are to help humans understand the world, they have to make sense of it and communicate about it in a way that's familiar and comprehensible to humans. This integration of data from various sensing technologies is beginning to happen in multimedia and visual analytics, where vision and sound are correlated. But that's just the start of what will be required in the next era of computing. (p.~74).

\end{quote}

\section{What would be better?}\label{what_would_be_better_section}

In \cite[Section 5]{marcus_2018}, Marcus writes: ``Despite all of the problems I have sketched, I don't think that we need to abandon deep learning. Rather, we need to reconceptualize it: not as a universal solvent, but simply as one tool among many, a power screwdriver in a world in which we also need hammers, wrenches, and pliers, not to mentions chisels and drills, voltmeters, logic probes, and oscilloscopes.'' (p.~18).

Yes, of course, in the spirit of maintaining several paths through the search space, it would be wrong to abandon all research on deep learning. But there is certainly a need to open up other areas and I believe the SP framework is one of them.

Regarding Marcus's remarks about symbolic and sub-symbolic systems \cite[Section 5.2]{marcus_2018}, I believe the SP system bridges that divide. In principle, it may work at any level of granularity.

In connection with this: ``The power and flexibility of the brain comes in part from its capacity to dynamically integrate many different computations in real-time. The process of scene perception, for instance, seamlessly integrates direct sensory information with complex abstractions about objects and their properties, lighting sources, and so forth.'' (p.~20), a major strength of the SP system, due largely to the powerful concept of SP-multiple-alignment, is the ability of the system to integrate diverse kinds of knowledge and diverse aspects of intelligence, in any combination.

As may be seen from Section \ref{psychology_neuroscience_section}, I agree very much that we should build ``models that are motivated not just by mathematics but also by clues from the strengths of human psychology.'' (p.~21).

\section{Conclusion}

The gist of this paper is that, while research on deep learning should certainly continue, the SP programme of research also merits attention. In that connection, the SP system has several advantages compared with deep learning systems. The main ones are:

\begin{itemize}

    \item {\em Quantities of data and one-trial learning (Section \ref{data_hungry_section})}. By contrast with deep learning systems, the SP system can produce meaningful results with quite small amounts of data. It provides a model for the way in which people can learn from a single exposure or experience. At the same time it provides an explanation for why it takes time to learn complex skills.

    \item {\em Hierarchical and non-hierarchical structures (Section \ref{problems_with_hierarchical_structure_section})}. Unlike deep learning systems---which do no lend themselves well to the representation of hierarchical structures---the SP system, via the concept of SP-multiple-alignment, accommodates such structures very well. At the same time, it also provides for the representation of non-hierarchical structures.

    \item {\em Reasoning (Sections \ref{open_ended_inference_section} and \ref{not_well_integrated_with_prior_knowledge_section})}. Although no attempt has yet been made to explore whether or how the SP system may perform `open ended' inference, the SP system has---unlike deep learning systems---strengths in several different kinds of reasoning. It also has strengths in `commonsense' reasoning, as described in \cite{sp_csrk}.

    \item {\em Transparency (Section \ref{transparency_section})}. By contrast with deep learning systems, the SP system provides complete transparency in the way in which it represents knowledge and it provides a full audit trail for all its processing.

    \item {\em It is unlikely that an SP system would be easily fooled (Section \ref{approximation_and_trust_section})}. Deep learning systems can misclassify stimuli in ways that people find eccentric or bizarre. Experience with the SP system, and a knowledge of how it works, suggests that it would be unlikely to make that kind of error.

    \item {\em Catastrophic forgetting (Section \ref{catastrophic_forgetting_section})}. A striking weakness of deep learning systems is `catastrophic forgetting'---the way in which new learning wipes out old learning. The SP system does not suffer from this problem because new learning does not disturb old learning.

    \item {\em Correction of over- and under-generalisations, and learning from `dirty data' (Section \ref{under-over-generalisation_section})}. With deep learning systems, a variety of solutions have been proposed for how to generalise correctly from a sample of data, without over- or under-generalisation, but it appears that none of them are entirely satisfactory. By contrast, the SP system proposes a theoretically-coherent solution that flows from the core theory in the system. Those same principles provide an explanation of how learning can be successful, despite errors in the data which is the basis for learning.

    \item {\em Psychology and neuroscience (Section \ref{psychology_neuroscience_section})}. By contrast with deep learning systems, which have long been recognised as being only loosely related to what is known about the workings of the human brain, the SP system draws extensively on research on human learning, perception, and cognition. The SP system also suggests how abstract structures and processes in the system may be realised in terms of neurons and their interconnections.

    \item {\em Overarching theory (Section \ref{overarching_theory_section})}. The central principle in the SP theory, derived from much research in cognitive psychology and supported by much evidence, is that much of human learning, perception, and cognition, may be understood as compression of information via the concept of SP-multiple-alignment. By contrast, deep learning systems have little or no over-arching theory.

\end{itemize}

In general, I believe that SP system provides a much firmer foundation than deep learning for the development of artificial general intelligence.

\bibliographystyle{plain}

\end{document}